\documentclass[letterpaper, 11 pt, conference]{ieeeconf}  % Comment this line out
\usepackage{graphicx} %package to manage images
\usepackage{multirow}
\usepackage{booktabs}
\usepackage[font=small]{caption}
\usepackage{subcaption}
\usepackage{array}
\usepackage{url}
\usepackage{hyperref}
\usepackage{xcolor}
\usepackage{cleveref}

\IEEEoverridecommandlockouts                              
\usepackage[utf8]{inputenc}
\usepackage[T1]{fontenc}

\title{\LARGE \bf Automatic prediction of mortality in patients with mental illness \\ using electronic health records}
\author{ Sean Kim$^{1, 2}$ and Samuel Kim$^{1}$ 
\\ \\
$^1$ IF Research Lab, La Palma, CA, USA \\
$^2$ Oxford Academy, Cypress, CA, USA \\
\\
{\tt\small seankim.hahjean@gmail.com} \\{\tt\small sam@ifresearchlab.com}
}

\begin{document}

\maketitle
\begin{abstract}
Mental disorders impact the lives of millions of people globally, not only impeding their day-to-day lives but also markedly reducing life expectancy. This paper addresses the persistent challenge of predicting mortality in patients with mental diagnoses using predictive machine-learning models with electronic health records (EHR). Data from patients with mental disease diagnoses were extracted from the well-known clinical MIMIC-III data set utilizing demographic, prescription, and procedural information. Four machine learning algorithms (Logistic Regression, Random Forest, Support Vector Machine, and K-Nearest Neighbors) were used, with results indicating that Random Forest and Support Vector Machine models outperformed others, with AUC scores of 0.911. Feature importance analysis revealed that drug prescriptions, particularly Morphine Sulfate, play a pivotal role in prediction. 
We applied a variety of machine learning algorithms to predict 30-day mortality followed by feature importance analysis. This study can be used to assist hospital workers in identifying at-risk patients to reduce excess mortality. 
\end{abstract}

\section{INTRODUCTION}
Mental disorders are a significant problem affecting 1 in 8 people worldwide, or around 970 million people~\cite{WHO}. Not only are mental disorders common, but they can also significantly affect patients' lives. Researchers have consistently found that patients with mental diagnoses are likelier to have a lower life expectancy~\cite{Hjorthøj2017, Olfson2015, Laursen2013, Liu2017, Laursen2011}. There are a variety of unique factors that contribute to reduced life expectancy. Unnatural causes like suicide and accidents accounted for 14 percent of excess mortality, while natural causes accounted for 86 percent~\cite{Mooij2019}. Accidental deaths were twice as common as suicide~\cite{Olfson2015}.  Various causes are associated with premature death, such as cardiovascular disease and substance-induced death, primarily from alcohol or drugs~\cite{Olfson2015, Mooij2019, Banerjee2021}. For example, diabetes, an extensive history of tobacco use, and a diagnosis of delirium were also found to be risk factors~\cite{Mooij2019, Brown2010, Banerjee2021}. Despite the numerous studies conducted to analyze factors of excess mortality in patients with mental diagnoses, the problem remains an ongoing concern. 

One popular data set used for mortality prediction is the MIMIC-III data set, a freely accessible, containing clinical data of patients admitted to critical care units~\cite{Johnson2016}. The data set allows clinical studies to be reproduced and serve as a basis for collaborative research. Being widely accessible to researchers, the data set has been used for various machine-learning applications in clinical settings, such as predicting hospital length of stay, predicting sepsis, and acute kidney injury prediction~\cite{Scherpf2019, Gentimis2017, Shawwa2020}. Regarding mortality prediction, the data set has been used to analyze mortality for numerous sub-patient groups, including patients with sepsis-3, heart failures, acute respiratory distress syndrome, or acute pancreatitis~\cite{Hou2020, Li2021, Ding2021, Huang2021}. However, literature examining mortality prediction among patients with mental diseases remains scarce. This paper aims to address this gap in existing research by utilizing a machine-learning-based approach to mortality prediction for patients with mental diagnoses with MIMIC-III. 

\section{Procedure}
\subsection{Data Source}
This study used the MIMIC-III (Medical Information Mart for Intensive Care) data set, which contains information on patients admitted to critical care units at Beth Israel Deaconess Medical Center in Boston, Massachusetts. The data includes 53,423 distinct hospital admissions for adult patients admitted between 2001 and 2012~\cite{Johnson2016}.

\subsection{Cohort definition}
From the data set, 13,400 patients with at least one mental disease diagnosis in their medical history were identified using the ICD-9 international classification of the diseases code system. From this group, 10,000 patients only had one hospital admission. 3,400 patients had at least one hospital admission, with number of admissions ranging from 2 to 42. For patients with more than one admission, only the data from the earliest hospital admission diagnosed with a mental disease was kept, while the other admissions were discarded. Thus, the final data set used in the study contains data on 13,400 unique patients and 13,400 unique admissions, where each patient only has one hospital admission. Data regarding the date of death was used in conjunction with discharge time to calculate mortality within 30 days. 1849 patients died (13.8\%), and 11551 patients (86.2\%) survived within 30 days. 
\subsection{Explanatory Data Analysis}
\begin{figure}
    \centering
    \includegraphics[width=1\linewidth]{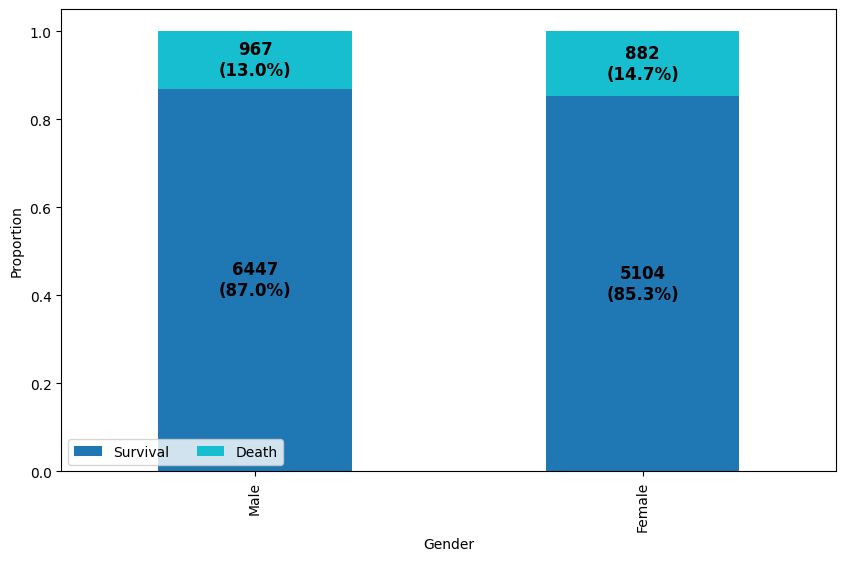}
    \caption{Mortality Distributions by Gender}
    \label{fig:gender}
\end{figure}

\begin{figure}
    \centering
    \includegraphics[width=1\linewidth]{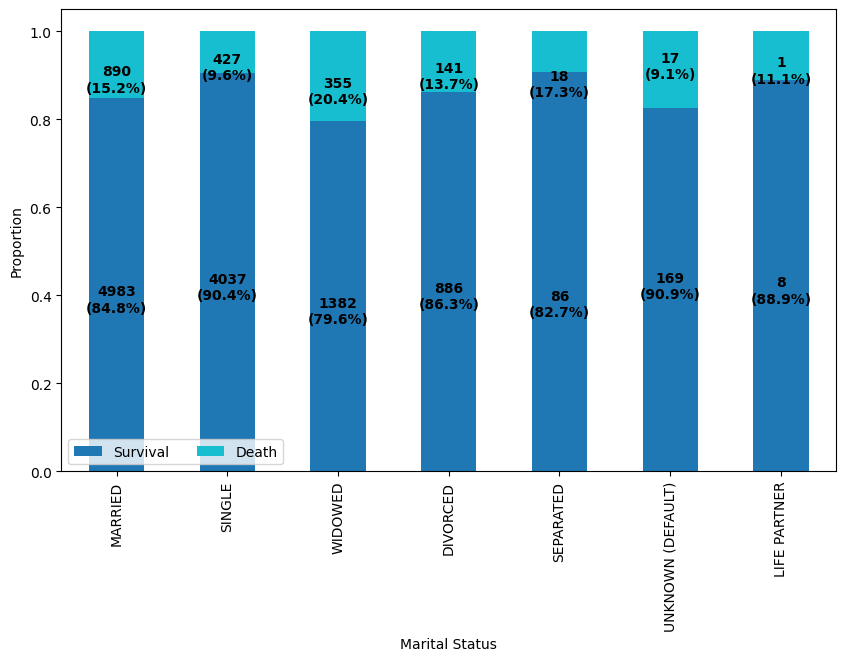}
    \caption{Mortality Distributions by Marital Status}
    \label{fig:marital}
\end{figure}

\begin{figure}
    \centering
    \includegraphics[width=1\linewidth]{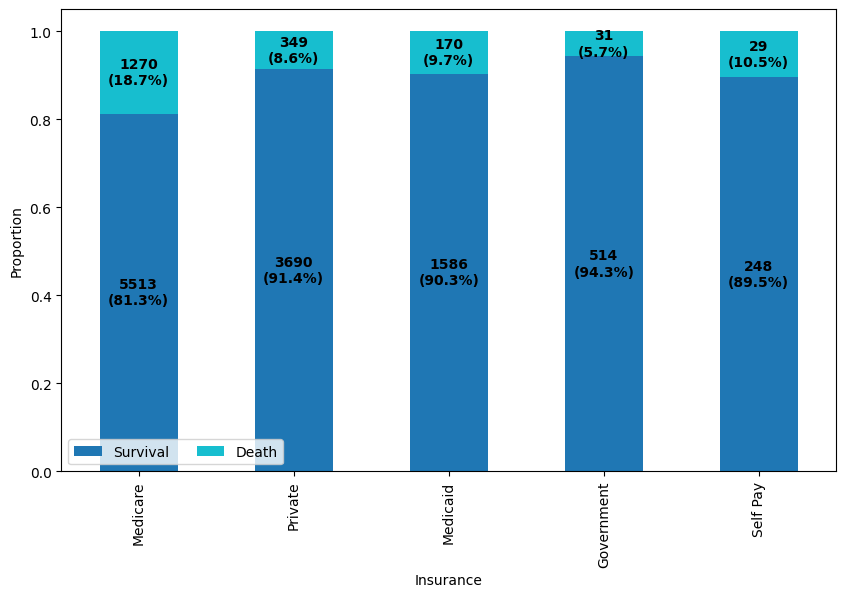}
    \caption{Mortality Distributions by Insurance}
    \label{fig:insurance}
\end{figure}

\begin{figure}
    \centering
    \includegraphics[width=1\linewidth]{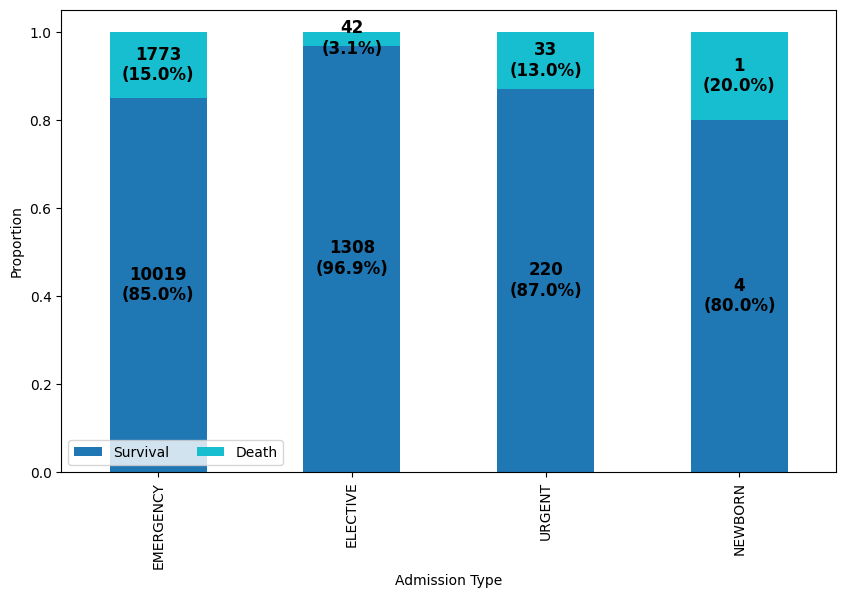}
    \caption{Mortality Distributions by Admission Type}
    \label{fig:admission}
\end{figure}

\begin{figure}
    \centering
    \includegraphics[width=1\linewidth]{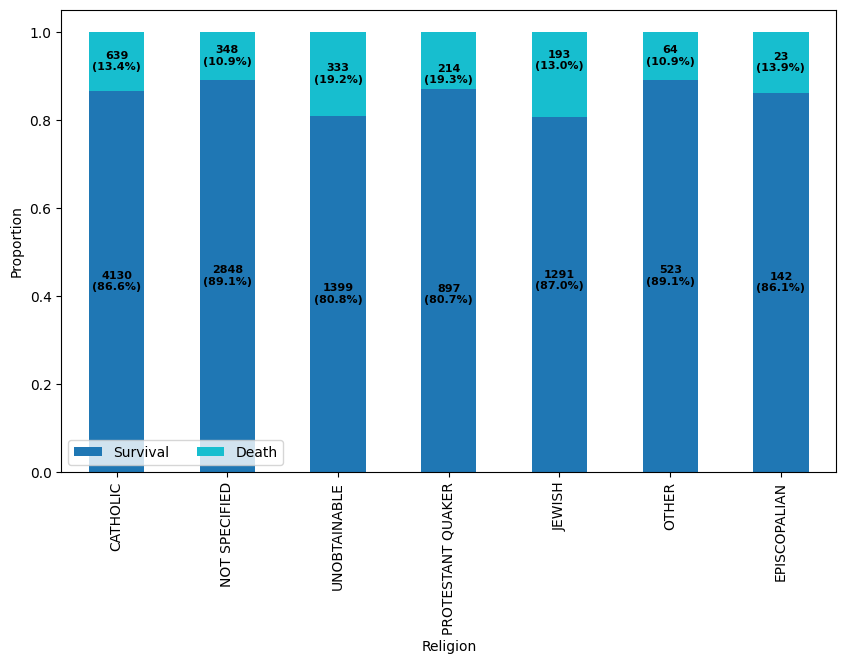}
    \caption{Mortality Distributions by Religion}
    \label{fig:religion}
\end{figure}

\begin{figure}
    \centering
    \includegraphics[width=1\linewidth]{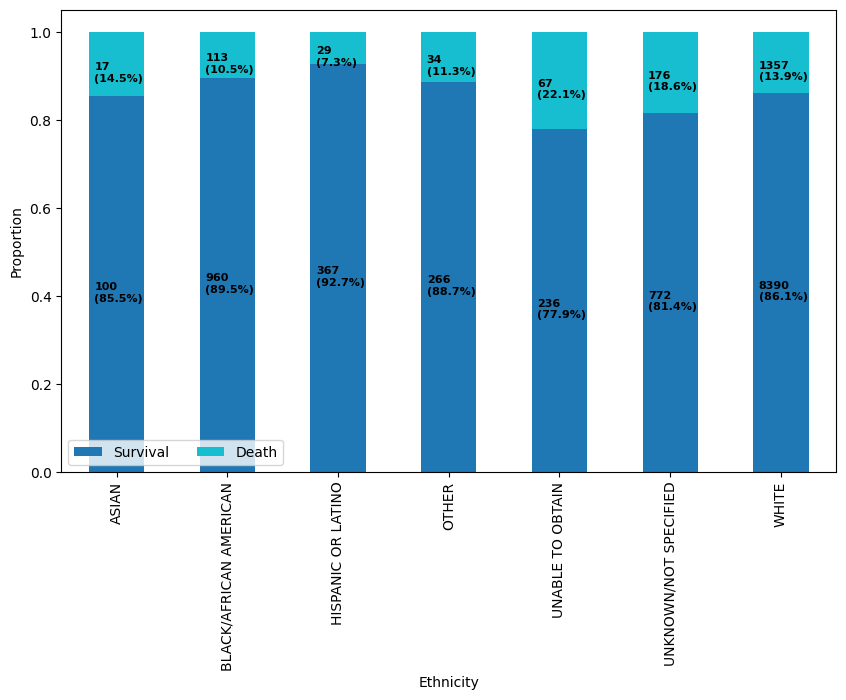}
    \caption{Mortality Distributions by Ethnicity}
    \label{fig:ethnicity}
\end{figure}

\begin{figure}
    \centering
    \includegraphics[width=1\linewidth]{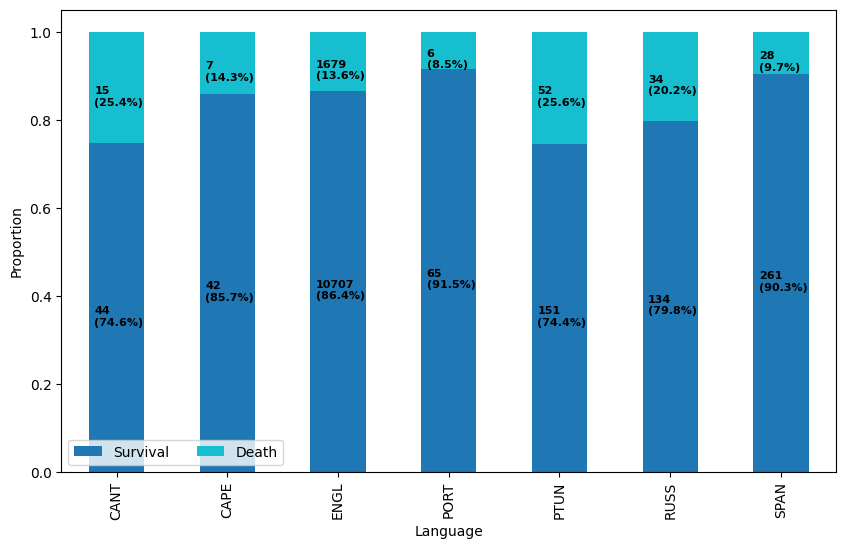}
    \caption{Mortality Distributions by Language}
    \label{fig:language}
\end{figure}

The distribution of mortality across seven different features (Gender, Marital Status, Insurance, Admission Type, Religion, Ethnicity, and Language) are shown below in \Cref{fig:gender,fig:marital,fig:insurance,fig:admission,fig:religion,fig:ethnicity,fig:language}. Within each graph, each bar is sorted from left to right by the number of deaths. \Cref{fig:religion,fig:ethnicity,fig:language} depicts the top seven most frequent categories within their respective features with respect to the number of deaths. 

\section{Experiments}
\subsection{Methodology}
This study primarily examined the impact of demographic, prescription, and procedural information in 30-day mortality prediction. Demographic data analyzed include language, marital status, religion, insurance, and ethnicity. Null demographic data was transformed using a simple imputer using a most frequent strategy. Prescription information was obtained by examining the Generic Sequence Number (GSN) of prescribed medication, with any null values being dropped. Procedural information was obtained from the ICD-9 Procedure Codes, with no null values found. Only prescription and procedural information from the included hospital admission was used. Demographic data was one-hot encoded, while prescription and procedural data utilized the Multi-Label Binarizer.

Four machine-learning algorithms were used to evaluate the performance of models in this task: Logistic Regression, Random Forest Classifier, Support Vector Machine (kernel=RBF, c=1), and K-Nearest Neighbors (k=5).  

\subsection{Results}

\begin{figure}[t!]
  \centering
      \includegraphics[width=8cm]{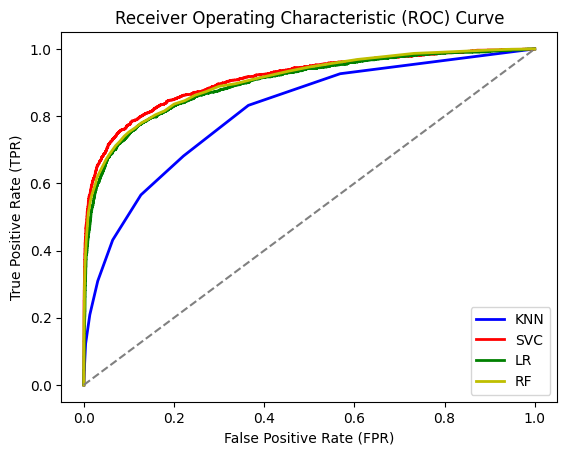}
  \caption{Receiver Operating Characteristic Curves of Algorithms}
  \label{roc}
\end{figure}

K-fold cross-validation was used to evaluate the model's performance (k = 5), with data split into five different shuffled folds. Fig.~\ref{roc} graphs the Receiver Operating Characteristic (ROC) curves for the Logistic Regression, Random Forest Classifier, Support Vector Machine, and K-Nearest Neighbor algorithms. Table~\ref{table:auc score} shows the area under the ROC curve (AUC) values for all four models. Based on the AUC score, the performance was 0.898 for Logistic Regression, 0.911 for Random Forest, 0.911 for Support Vector Machine, and 0.812 for K-Nearest Neighbor. The Random Forest and Support Vector classifiers resulted in the highest performance.

\begin{table}[t!]
\centering

\begin{tabular}{c|c|c|c|c}
\hline 
 & RF & SVC & LR & KNN \\
\hline \hline
\multirow{1}{*}{ROC-AUC} & 0.911 & 0.911 & 0.898 & 0.812 \\
\hline
\end{tabular}
 \caption{Average ROC-AUC scores with various machine learning algorithms.}
\label{table:auc score}
\end{table}

\subsection{Feature Importance}
Feature importance values were calculated to obtain insight into the importance of input features for the Random Forest Classifier. Table ~\ref{tab:feature-importance} lists the top ten features and feature importance scores calculated using Permutation Importance. The top three most critical features were all types of drugs: Morphine Sulfate, Scopolamine Patch, and Pantoprazole, respectively. Of the top ten features, nine were drug prescriptions, only one was a procedure (Continuous invasive mechanical ventilation for less than 96 consecutive hours), and no demographic factors were present. This shows that prescription information seems to be the most relevant when predicting mortality for patients with mental illnesses, with procedural and demographic data less relevant to model performance. 

\begin{table}[htbp]
  \centering
  \begin{tabular}{>{\centering\arraybackslash}m{5cm}|>{\centering\arraybackslash}m{2cm}}
    \hline
    \textbf{Features} & \textbf{Importance} \\
    \hline \hline
    Morphine Sulfate & 0.00258 \\
    \hline
    Scopolamine Patch & 0.00646 \\
    \hline
    Pantoprazole & 0.00415 \\
    \hline
    Oxycodone-Acetaminophen & 0.00415 \\
    \hline
    Docusate Sodium & 0.00369 \\
    \hline
    Heparin & 0.00369 \\
    \hline
    Vasopressin & 0.00369 \\
    \hline
    Fentanyl Citrate & 0.00323 \\
    \hline
    Continuous invasive mechanical ventilation for less than 96 consecutive hours
 & 0.3230 \\
    \hline
    Piperacillin-Tazobactam Na & 0.00277 \\
    \hline
  \end{tabular}
  \caption{Feature Importance for Random Forest Model}
  \label{tab:feature-importance}
\end{table}

\subsection{Discussion}
The Support Vector Machine had the highest performance of the models tested. The Support Vector Machine and Random Forest models performed similarly, with an AUC value of 0.911. Trailing was the Logistic Regression model at 0.898, then K Nearest Neighbor with an AUC value of 0.812.
Morphine Sulfate was found to have the highest importance, possibly due to its role in severe pain relief. It may be correlated with mortality due to the fact that higher-risk patients require higher levels of pain relief. A causal study into causes of mortality in mental disease patients may be a topic of interest for future work. 

\subsection{Limitations}
A limitation of this study is that although this study looked at mental disorders as a broad category, illnesses vary widely in severity; certain groups of more severe disorders may result in more deadly symptoms than mild mental disorders. These differences across severity may heavily impact the mortality rates. To address this concern, future research can consider the finer distinctions between severity among mental disease diagnoses. 

\section{Conclusion}
This study used machine-learning-based approaches to tackle the problem of mortality prediction for patients diagnosed with mental illnesses using MIMIC-3 data. Support Vector Classifier proved to have the highest performance of the examined machine learning classifiers. 
This research can be applied in a hospital setting to identify at-risk mental illness patients at an early stage of their hospital stay and medical journey to reduce premature mortality. 

\bibliographystyle{IEEEbib}
{\bibliography{mybib}}

\end{document}